\documentclass[letterpaper,journal]{IEEEtran}
\IEEEoverridecommandlockouts
\def\BibTeX{{\rm B\kern-.05em{\sc i\kern-.025em b}\kern-.08em
    T\kern-.1667em\lower.7ex\hbox{E}\kern-.125emX}}
\usepackage{balance}

\usepackage{graphicx}
\usepackage{xcolor}
\usepackage{algorithm}

\usepackage[hidelinks]{hyperref}
\usepackage{multirow}
\usepackage{booktabs}
\usepackage{makecell}
\usepackage{algpseudocode} 
\usepackage{optidef}
\usepackage{mathtools}

\usepackage{amsfonts,amssymb,amsthm}
\usepackage{tabularray} 
\DeclareGraphicsExtensions{.pdf,.png,.jpg}

\usepackage{tabularx}

\usepackage{array}

\usepackage[caption=false,font=normalsize,labelfont=sf,textfont=sf]{subfig}

\usepackage{url}

\usepackage[noadjust]{cite}

\begin{document}

\newcommand*{\vertbar}{\rule[-1ex]{0.5pt}{2.5ex}}
\newcommand*{\horzbar}{\rule[.5ex]{2.5ex}{0.5pt}}
\newcommand{\di}[1]{\mathrm{d}#1}
\title{Controlling Continuum Robot Shape in Enclosed Space with Embedded Sensing using IMU and Actuator Feedback}





\author{
Jiahe Wang$^{\dagger}$, Student Member, IEEE,
Sultan Haidar Ali$^{\dagger}$,
Eric Weissman, Student Member, IEEE,
\\ and Jiefeng Sun*, Member, IEEE%
\thanks{$^{\dagger}$Jiahe Wang and Sultan Haidar Ali contributed equally to this work. All authors are with the School for Engineering of Matter, Transport and Energy, Arizona State University, Tempe, AZ 85281, USA (email: {\tt\small jwang810@asu.edu, sultanha@asu.edu, eweissm1@asu.edu, jiefeng.sun@asu.edu}).}
\thanks{This work was supported in part by the Ira A. Fulton Schools of Engineering at Arizona State University. \textit{(Corresponding author: Jiefeng Sun)}}
\thanks{GitHub: \href{https://github.com/sunroboticslab/Embedded_Shape_Sensing_Tmech2027}{\textcolor{blue}{github.com/sunroboticslab/Embedded\_Shape\_Sensing\_Tmech2027}}. Additional experimental data are available from the first author upon reasonable request.}

}

\maketitle

\begin{abstract}
Controlling continuum robots' shape to operate in confined, enclosed environments is challenging, since it cannot rely on external observation. Existing embedded shape sensing methods, although they can provide shape reconstruction, are insufficient to enable real-time shape control due to limitations in reconstruction accuracy, computational speed, and compatibility with shape-state representation.
In this work, we present an embedded shape-sensing framework that uses readily available IMU and actuator feedback to reconstruct the continuum-robot shape with a neural network, providing accurate shape feedback at $400~\mathrm{Hz}$ for physical closed-loop control. We systematically investigate four key reconstruction design choices: shape-state representation, network architecture, training objective, and sensing inputs, through more than $5{,}000$ trained models on one- and three-segment continuum robots. The selected framework achieves global shape errors of $0.9\%L$ and $1.7\%L$ (of the robot length) for the one- and three-segment systems, respectively. In closed-loop control, one-segment tip tracking achieves a $6.1\%L$ error and three-segment shape control achieves a $2.3\%L$ global shape error, whereas IMU--PCC reconstruction using the same uncompensated measurements fails. Enclosed-space experiments further demonstrate operation without external observation. These results provide a practical framework for translating readily available embedded sensing into high-rate feedback for continuum robot closed-loop control.
\end{abstract}



\section{Introduction}


Continuum robots achieve compliant motion through distributed deformation, adapting their bodies to unstructured environments and operating in enclosed spaces \cite{mielke_control_2026}. However, existing continuum-robot operation in enclosed or visually occluded environments remains challenging \cite{li_autonomous_2025,qing_minimal-actuation_2026}. The major limitation is that the shape state of continuum robots' feedback often relies on external observation, such as motion capture, which requires an instrumented workspace and continuous line of sight~\cite{teetaert_continuum_2026}. Embedded shape sensing, reconstructing the shape state from measurements available on the robot itself, is the most promising option.


Existing embedded shape sensing has primarily focused on improving reconstruction accuracy, with comparatively limited consideration of downstream closed-loop control, as summarized in Tab.~\ref{tab:shape_sensing_comparison}. First, achieving low reconstruction error and a high update rate for real-time closed-loop control is challenging. Distributed IMUs~\cite{stella_soft_2024, peng_tendon-driven_2024, peiIMUBasedPose2024a}, magnetic sensors~\cite{adamu_3-axis_2025}, and strain sensors~\cite{feliu-talegon_advancing_2025,wang_sensing_2024,wang_strain-based_2024} provide local orientation or deformation measurements that can be combined with continuum kinematics to recover the robot shape. Accurate reconstruction requires sensor-specific calibration and processing, while piecewise-constant-curvature (PCC) assumptions may not hold under nonuniform bending. Optical-fiber-based sensing can provide accurate distributed deformation measurements, but often requires dedicated interrogation hardware, complex fabrication, and limited sensing lengths \cite{julia_wise_enabling_2026,he_deep_2026,tang_soft_2024}. Actuator feedback, such as motor encoder and torque measurements, provides a more readily available sensing source \cite{feliu-talegon_actuation_2025,feliu-talegon_dynamic_2025}, but only indirectly reflects the robot configuration. Recovering the shape from these measurements through physical models introduces additional iterative estimation and computational requirements \cite{zheng_estimating_2026}.

\begin{figure}[t]
    \centering    
    \includegraphics[width=1\linewidth]{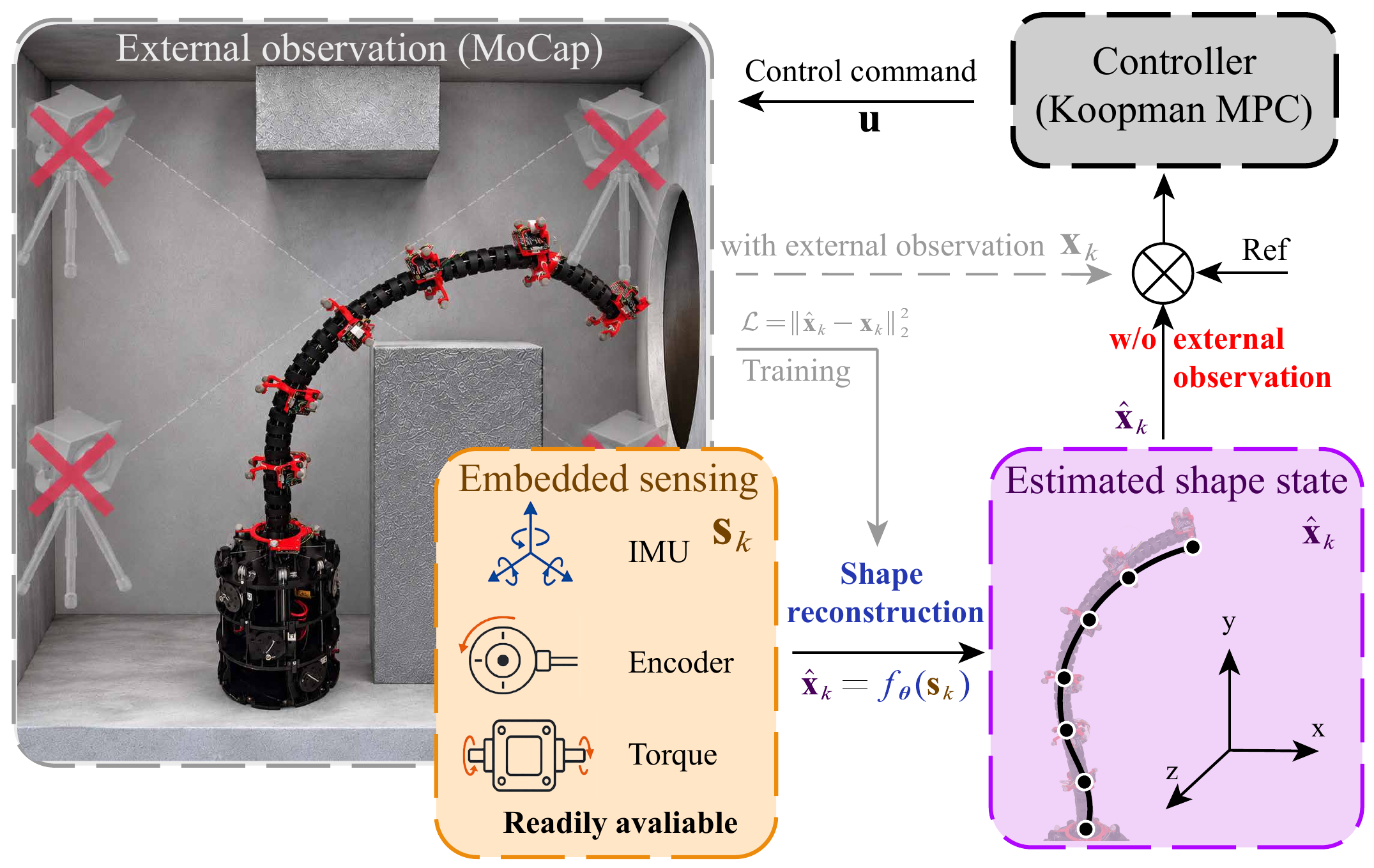}
    \caption{Overview of the proposed embedded shape-sensing and control framework for continuum robots working in enclosed space. Readily available embedded sensing reconstructs the shape state $\hat{\mathbf{x}}_k$ for closed-loop control without external observation. External motion capture provides ground-truth supervision only during training. Video: \href{https://youtu.be/sBmrXCk2B-4}{\textcolor{blue}{https://youtu.be/sBmrXCk2B-4}}.}
    \label{fig:title_fig}
\end{figure}

Second, even when relatively low reconstruction error and a high update rate can be achieved, a gap remains between the reconstructed state and the closed-loop control state. Learning-based approaches directly map embedded measurements to a selected representation of the reconstructed state, for example, by combining actuator feedback with machine learning \cite{yong_location_2026,goh_3d_2026,zhan_physics_2025} or by combining optical sensing with deep learning \cite{he_deep_2026, hu_stretchable_2023}. However, the choice of reconstructed shape state affects both the accuracy of the estimation and the geometric information available for control. Including orientation retains richer geometric information but introduces higher-dimensional and constrained variables that are more difficult to reconstruct accurately. Simpler position-based representations may omit segment-local deformation information relevant to full-body shape control \cite{wang_real-time_2026}. Beyond the shape-state representation, network architecture, training objective, and sensing inputs can further affect reconstruction accuracy, temporal consistency, and computational cost. The key challenge is therefore not only to reconstruct the continuum-robot shape state accurately and efficiently, but to systematically design a control-oriented reconstruction framework.


\begin{table}[t]
\centering
\caption{Comparison of embedded shape-sensing methods and their validation in closed-loop continuum-robot control.}
\label{tab:shape_sensing_comparison}
\renewcommand{\arraystretch}{1.08}
\setlength{\tabcolsep}{1.5pt}
\scriptsize

\begin{tabularx}{\columnwidth}{
    >{\raggedright\arraybackslash}p{0.075\columnwidth}
    >{\centering\arraybackslash}p{0.045\columnwidth}
    >{\centering\arraybackslash}p{0.245\columnwidth}
    >{\centering\arraybackslash}X
    >{\centering\arraybackslash}p{0.09\columnwidth}
    >{\centering\arraybackslash}p{0.20\columnwidth}
}
\toprule
\textbf{Ref.} &
\textbf{Seg.} &
\textbf{Sensor} &
\makecell{\textbf{Reconstructed}\\\textbf{Tip Error (\%$L$)}} &
\makecell{\textbf{Rate}\\\textbf{(Hz)}} &
\makecell{\textbf{Closed-Loop}\\\textbf{Control}} \\
\midrule

\cite{cheng_orientation_2022}
& 1 & IMU & 0.91 & 60 & $\times$ \\
& 2 & IMU & 1.21 & 60 & $\times$ \\

\cite{wang_sensing_2024}
& 2 & Strain & 1.8 & -- & $\checkmark$ (tip) \\

\cite{feliu-talegon_advancing_2025}
& 3 & Strain & 2.6 & -- & $\times$ \\

\cite{an_shape_2024}
& 2 & \makecell{FBG + IMU} & 3.6 & -- & $\times$ \\

\cite{stella_soft_2024}
& 2 & IMU & 4.1 & 20 & $\times$ \\

\cite{peng_tendon-driven_2024}
& 3 & IMU & 4.4 & -- & $\times$ \\

\cite{adamu_3-axis_2025}
& 2 & Magnetic & 4.8 & 90 & $\times$ \\

\cite{feliu-talegon_dynamic_2025}
& 2 & Actuator feedback & 7.2 & -- & $\times$ \\

\midrule

\multirow{2}{*}{\makecell{\textbf{Our}\\\textbf{work}}}
& 1 & \multirow{2}{*}{\makecell{\textbf{IMU}\textbf{+ actuator feedback}}}
& \textbf{1.6}
& \multirow{2}{*}{\textbf{400}}
& $\checkmark$ \textbf{(tip)} \\

& 3 & 
& \textbf{1.8}
&
& $\checkmark$ \textbf{(shape)} \\

\bottomrule

\multicolumn{6}{p{0.97\columnwidth}}{\scriptsize
\textbf{Note:} $L$ denotes robot length; FBG denotes fiber Bragg grating; ``--'' indicates not reported; $\times$ indicates that closed-loop control using the reconstructed state was not demonstrated.}
\end{tabularx}
\vspace{-10pt}
\end{table}


Consequently, as summarized in Tab.~\ref{tab:shape_sensing_comparison}, a gap remains between embedded shape reconstruction and its systematic validation on closed-loop control for continuum robots. This challenge arises because these requirements (real-time computation, reconstruction accuracy, and representation compatibility) depend on multiple reconstruction design choices, including shape-state representation, sensing inputs, network architecture, and training objective. For example, increasing model complexity or temporal history may improve reconstruction accuracy but increase computation and reduce the achievable feedback rate. Despite these considerations, existing studies have primarily evaluated embedded shape sensing through reconstruction metrics, with comparatively limited task-level control validation \cite{yong_location_2026,zhan_physics_2025,kasaei_synergistic_2025,cheng_orientation_2022}.


In this work, we develop a practical embedded shape-sensing and control framework that enables accurate and robust shape control of a continuum robot in an enclosed environment. 
The framework reconstructs the robot shape from readily available IMU, motor encoder, and torque measurements without requiring a dedicated sensor-specific filtering or drift-compensation pipeline, simplifying the implementation of embedded shape sensing. We then systematically investigate four key reconstruction design choices: shape-state representation, network architecture, training objective, and sensing inputs for neural network (NN) training. The resulting framework operates at $400~\mathrm{Hz}$, with the reconstructed shape state used directly for Koopman identification and model predictive control (MPC) \cite{bruderDataDrivenControlSoft2021c,wang_real-time_2026,haggertyControlSoftRobots2023a}. The framework is experimentally validated through one-segment tip tracking, three-segment shape control, and enclosed-space operation without external observation. The main contributions are as follows:
\begin{enumerate}
    \item An embedded shape sensing and control framework for continuum-robot closed-loop control using readily available IMUs and actuator feedback, without dedicated sensor-specific filtering or drift compensation;

    \item A systematic investigation of the key factors influencing shape reconstruction performance, including network architecture, shape-state representation, training objective, and sensing inputs, through more than $5{,}000$ trained models on one- and three-segment continuum robots;

    \item Task-level control validation beyond modeling accuracy, demonstrating that the reconstructed state can serve directly as high-rate feedback for physical closed-loop control, including one-segment tip tracking, three-segment shape control, and enclosed-space operation without external observation.
\end{enumerate}

The remainder of this paper is organized as follows. Sec.~\ref{sec:framework} formulates the embedded shape-sensing and control framework. Sec.~\ref{sec:methods} presents the physical implementation, model training, and reconstruction design study. Sec.~\ref{sec:experiment} evaluates shape reconstruction and closed-loop control, including enclosed-space demonstrations. Sec.~\ref{sec:conclusion} concludes the paper.

\section{Embedded Shape Sensing and Control Framework}
\label{sec:framework}

The objective is to estimate the shape of a continuum robot from embedded measurements with sufficient accuracy and update rate for closed-loop control. As shown in Fig.~\ref{fig:title_fig}, $\mathbf{x}_k$ denotes the ground-truth shape state of the continuum robot at time step $k$; $\mathbf{s}_k$ denotes the embedded sensing measurements; $\hat{\mathbf{x}}_k$ denotes the reconstructed shape state. The reconstruction is represented by $\hat{\mathbf{x}}_k=f_{\boldsymbol{\theta}}(\mathbf{s}_{k-H+1:k})$, where $f_{\boldsymbol{\theta}}$ denotes the reconstruction model with parameters $\boldsymbol{\theta}$ and $H$ denotes the measurement history. During offline training, $\boldsymbol{\theta}$ is identified by minimizing a training objective $\mathcal{L}(\hat{\mathbf{x}}_k,\mathbf{x}_k)$.

The resulting reconstruction depends on four coupled design choices: the representation of $\mathbf{x}_k$, the architecture of $f_{\boldsymbol{\theta}}$, the training objective $\mathcal{L}$, and the measurements included in $\mathbf{s}_k$. We introduce these four components below and subsequently describe how the reconstructed state $\hat{\mathbf{x}}_k$ is used for Koopman-based modeling and closed-loop control.

\subsection{Shape-State Representation}
\label{subsec:shape_state}

The first design choice is the representation of the reconstructed shape state $\mathbf{x}_k$. A richer state retains more geometric information but increases the reconstruction dimension and may introduce variables that are difficult to estimate accurately. Consider a continuum robot represented by $N_b$ discrete body frames distributed along its backbone. Let $\{F\}$ denote the frame attached to the robot base, and let $\{B_j\}$ denote the local base frame of segment $j$. The position and orientation of the $i$-th body frame expressed in $\{F\}$ are denoted by $\mathbf{p}_i^{\{F\}}\in\mathbb{R}^{3}$ and $\mathbf{q}_i^{\{F\}}\in\mathbb{S}^{3}$, respectively. In this paper, we introduce the following three shape-state representations.

The first representation directly stacks the positions and orientations represented under the world (global) frame along the robot,
\begin{equation}
\mathbf{x}^{pq}
=
\left[
    \mathbf{p}_{2}^{\{F\}\top},
    \mathbf{q}_{2}^{\{F\}\top},
    \ldots,
    \mathbf{p}_{N_b}^{\{F\}\top},
    \mathbf{q}_{N_b}^{\{F\}\top}
\right]^{\top}.
\label{eq:state_pq}
\end{equation}

This representation has been widely used in soft-robot \cite{chenVersatileNeuralNetwork2025b,haggertyControlSoftRobots2023a} and is closely related to IMU-based reconstruction methods, where distributed orientation measurements are used to recover the continuum-robot configuration \cite{stella_soft_2024,peng_tendon-driven_2024,cheng_orientation_2022}. However, including orientation increases the state dimension and introduces constrained geometric variables. For example, quaternion states must satisfy unit-norm constraints. 

Alternatively, the robot shape can be represented using only positions while retaining both global and segment-local geometric information \cite{ristich_shape_2025}. Let $\mathbf{p}_i^{\{B_{s(i)}\}}$ denote the position of body frame $i$ expressed relative to the base frame of its associated segment $s(i)$. We define
\begin{equation}
\mathbf{x}^{pp}
=
\left[
    \mathbf{p}_{2}^{\{F\}\top},
    \ldots,
    \mathbf{p}_{N_b}^{\{F\}\top},
    \mathbf{p}_{2}^{\{B_{s(2)}\}\top},
    \ldots,
    \mathbf{p}_{N_b}^{\{B_{s(N_b)}\}\top}
\right]^{\top}, 
\label{eq:state_pp}
\end{equation}
 where the global positions describe the overall robot configuration, whereas the local positions explicitly encode the deformation of each segment.  This representation avoids explicit orientation states and is well suited to continuum robots with limited axial elongation, where deformation is dominated by bending, following the global--local projection strategy for multi-segment shape control \cite{wang_real-time_2026}.

Finally, a truncated representation can be obtained by retaining only the coordinates in the plane normal to the nominal backbone direction. For the coordinate convention used in this work, the $y$-axis is aligned with the nominal backbone, and the $x$-$z$ plane captures the dominant bending deformation. Defining
\begin{equation}
\mathbf{p}_{i,xz}^{\{\cdot\}}
=
\begin{bmatrix}
1 & 0 & 0\\
0 & 0 & 1
\end{bmatrix}
\mathbf{p}_{i}^{\{\cdot\}},
\label{eq:xz_projection}
\end{equation}
the truncated global--local state is
\begin{equation}
\mathbf{x}^{pp}_{xz}
=
\left[
    \mathbf{p}_{2,xz}^{\{F\}\top},
    \ldots,
    \mathbf{p}_{N_b,xz}^{\{F\}\top},
    \mathbf{p}_{2,xz}^{\{B_{s(2)}\}\top},
    \ldots,
    \mathbf{p}_{N_b,xz}^{\{B_{s(N_b)}\}\top}
\right]^{\top}.
\label{eq:state_pp_xz}
\end{equation}

Removing the extension direction reduces the state dimension and can mitigate modeling errors associated with the highly nonlinear axial direction, as observed in \cite{wang_real-time_2026}. 

 Eq.~\eqref{eq:state_pq}--\eqref{eq:state_pp_xz} provide different representations of geometric information. Sparser shape states were also evaluated, providing lower computational cost and lower reconstruction accuracy. These results are omitted for brevity.

\subsection{Reconstruction Models Architecture}
\label{subsec:network_architecture}

The second design choice is the reconstruction model $f_{\boldsymbol{\theta}}$. We consider a multilayer perceptron (MLP) as a one-to-one mapping from the current embedded measurements to the shape state, and a long short-term memory (LSTM) to incorporate temporal information through a measurement history. The MLP provides a simple and computationally efficient estimator, but its instantaneous one-to-one mapping can directly propagate measurement noise into the reconstructed state. In contrast, the LSTM exploits temporal information that may improve consistency of the reconstructed state, at the cost of increased computation and latency.

Temporal convolutional networks (TCNs) were also evaluated as an additional temporal architecture; however, they did not outperform the selected MLP or LSTM in this paper, and are omitted from the subsequent comparison for brevity.

\subsection{Reconstruction Model Training Objective}
\label{subsec:training_objective}

The third design choice is the training objective $\mathcal{L}$ used to identify $\boldsymbol{\theta}$. A standard choice is the mean-squared error (MSE) between the reconstructed and ground-truth shape states,
\begin{equation}
    \mathcal{L}_{\mathrm{MSE}}
    =
    \frac{1}{N_D}
    \sum_{k=1}^{N_D}
    \left\|
        \hat{\mathbf{x}}_k-\mathbf{x}_k
    \right\|_2^2 ,
    \label{eq:mse_loss}
\end{equation}
where $N_D$ denotes the number of training samples.

For multi-segment shape reconstruction, element-wise MSE does not explicitly distinguish overall configuration error from deformation within individual segments. We therefore additionally consider the global--local objective motivated by \cite{wang_real-time_2026},
\begin{equation}
    \mathcal{L}_{\mathrm{G}\_\mathrm{local}}
    =
    \alpha e_{\mathrm{global}}
    +(1-\alpha)e_{\mathrm{local}},
    \qquad \alpha\in[0,1],
    \label{eq:glocal_loss}
\end{equation}
where $e_{\mathrm{global}}$ and $e_{\mathrm{local}}$ denote the global and segment-local geometric reconstruction errors defined in Sec.~\ref{subsec:shape_error}.

\subsection{Embedded Sensing Inputs}
\label{subsec:sensing_inputs}

The fourth design choice is the composition of the embedded measurement vector $\mathbf{s}_k$. IMU, encoder, and torque are readily available from the system. To quantify the contribution of each sensing modality, we consider four representative configurations with progressively increasing information, as detailed in Fig.~\ref{fig:reconstruction_ablation}(d, h).

Additional sensing combinations and spatially sparser configurations, including fewer distributed IMUs, were also evaluated. These configurations generally resulted in higher reconstruction errors and are omitted for brevity. The collected dataset is publicly available on GitHub to facilitate further evaluation of alternative sensing configurations.

\subsection{Koopman Modeling and Control}
\label{subsec:koopman_control}

The reconstructed state $\hat{\mathbf{x}}_k=f_{\boldsymbol{\theta}}(\mathbf{s}_k)$ can be used with different downstream control approaches. Importantly, the proposed reconstruction framework is not restricted to a specific control approach.
In this work, we adopt Koopman-based model predictive control (K-MPC), which has demonstrated effective real-time control of continuum robots \cite{bruderDataDrivenControlSoft2021c,haggertyControlSoftRobots2023a,wang_real-time_2026}. To capture temporal dynamics, a delay-embedded state is constructed from the reconstructed shape,
$\mathbf{z}_k
    =
    \begin{bmatrix}
        \hat{\mathbf{x}}_k^\top &
        \hat{\mathbf{x}}_{k-1}^\top &
        \cdots &
        \hat{\mathbf{x}}_{k-d}^\top
    \end{bmatrix}^{\top},
$
where $d$ denotes the number of delays. The resulting dynamics are approximated by
\begin{equation}
    \mathbf{z}_{k+1}=A\mathbf{z}_k+B\mathbf{u}_k,
    \qquad
    \hat{\mathbf{x}}_k=C\mathbf{z}_k,
    \label{eq:koopman_model}
\end{equation}
where $A$ and $B$ are identified from trajectories of the reconstructed shape state and actuation input $\mathbf{u}_k$~\cite{kordaLinearPredictorsNonlinear2018}.

For control, a task-dependent output is defined as
$\mathbf{y}_k=C_{\mathrm{task}}\mathbf{z}_k$, where
$C_{\mathrm{task}}$ selects the tip position for tip tracking or
distributed body positions for full-body shape control. Given a reference
trajectory $\mathbf{r}_{k+i}$ over a prediction horizon $N_H$, the control
input is obtained by solving
\begin{equation}
\begin{aligned}
\min_{\mathbf{u}_{k:k+N_H-1}}
&\sum_{i=0}^{N_H-1}
\left(
\|\mathbf{y}_{k+i}-\mathbf{r}_{k+i}\|_Q^2
+
\|\mathbf{u}_{k+i}\|_R^2
\right) \\
&+
\|\mathbf{y}_{k+N_H}-\mathbf{r}_{k+N_H}\|_P^2 ,
\end{aligned}
\label{eq:kmpc}
\end{equation}
subject to the Koopman dynamics in Eq.~\eqref{eq:koopman_model} and actuator constraints \cite{kordaLinearPredictorsNonlinear2018}. The same reconstructed shape state is therefore used for both Koopman identification and online feedback, forming the sensing-to-control pipeline evaluated in Sec.~\ref{sec:experiment}.
\section{Hardware and Shape Reconstruction Analysis }
\label{sec:methods}

This section describes the physical implementation and analyzes the influence of model selections as introduced in Sec.~\ref{sec:framework}. We first introduce the robot hardware and collect synchronized embedded measurements and ground-truth shape data. The resulting dataset is then used to train the candidate reconstruction models, whose performance is evaluated using the proposed shape errors. Finally, a large-scale ablation study identifies the reconstruction configuration, and the selected NN is subsequently used for closed-loop control.

\subsection{Robot Hardware and Data Collection}
\label{subsec:data_collection}

\begin{figure}[t]
    \centering
    \includegraphics[width=0.75\linewidth]{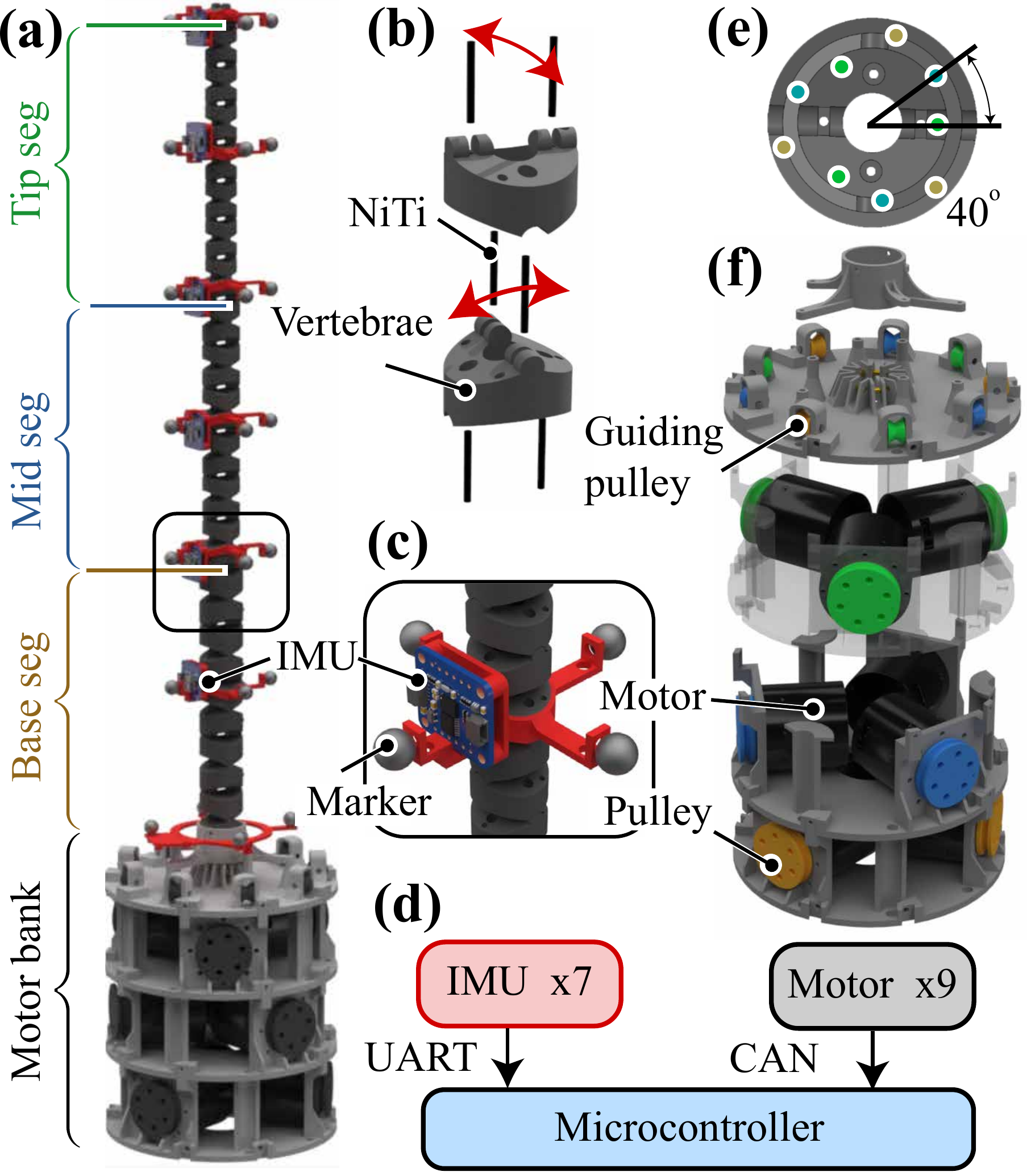}
    \caption{Three-segment continuum robot and embedded sensing system. (a) Physical prototype consisting of base, middle, and tip segments mounted on a three-layer motor bank. (b) Interlocking vertebrae and orthogonally arranged NiTi hinge. (c) Integrated IMU and motion-capture marker mount. (d) Communication architecture for seven IMUs and nine motors. (e) Tendon-routing configuration within each segment. (f) Three-layer motor bank with guiding pulleys and three motors per layer.}
    \label{fig:physcial_prototype}
\end{figure}

The three-segment continuum robot, with a total length of $516~\mathrm{mm}$ and a diameter of 20~mm, is shown in Fig.~\ref{fig:physcial_prototype}(a). Inspired by \cite{wang_design_2018}, each segment consists of a series of interlocking rigid vertebrae (SLA-printed in Tough 2000 V2 resin) connected by 1.2~mm-diameter NiTi rods, as shown in Fig.~\ref{fig:physcial_prototype}(b). An 8~mm central bore runs through the robot to accommodate the embedded sensor wiring.

Each segment is actuated by three tendons (PowerPro 100-lb microfilament braided line) uniformly distributed around the central axis and terminated at the distal end of the corresponding segment, as shown in Fig.~\ref{fig:physcial_prototype}(e). The tendons are routed through guiding pulleys to a three-layer motor bank, with each layer containing three quasi-direct-drive (QDD) motors (RobStride~05), as shown in Fig.~\ref{fig:physcial_prototype}(f). The motors are regulated by PD control with a constant feedforward torque to maintain tendon pretension and prevent slack.

For embedded shape sensing, seven calibrated IMUs (Adafruit BNO085) are distributed along the robot body and provide orientation measurements with no additional dedicated drift-compensation pipeline. Motor encoder positions and torque feedback are directly available from the actuation system. A Teensy~4.1 collects the IMU and actuator measurements at $400~\mathrm{Hz}$ and transmits them to a host computer running ROS~2. Four infrared markers are mounted on each plate for OptiTrack motion capture, as shown in Fig.~\ref{fig:physcial_prototype}(c), providing the ground-truth body poses required only during offline data collection.

During data collection, the command input $\mathbf{u}_k\in\mathbb{R}^{6}$ consists of two curvature commands per segment, which are mapped by a PCC model to nominal motor encoder commands, while compliant PD control with feedforward tendon pretension allows the actual motor positions and robot shape to deviate from the nominal PCC configuration. Ramp-and-hold input trajectories \cite{bruderDataDrivenControlSoft2021c} are used to explore the robot workspace. OptiTrack records the poses of seven distributed rigid bodies at $100~\mathrm{Hz}$, with each pose consisting of a 3D position and a quaternion, yielding the ground-truth state $\mathbf{x}_k\in\mathbb{R}^{49}$. The embedded measurement vector $\mathbf{s}_k\in\mathbb{R}^{46}$ consists of encoder and torque measurements from nine motors ($18$ states) and quaternion measurements from seven IMUs ($28$ states), acquired at $400~\mathrm{Hz}$ and temporally aligned with the motion-capture measurements.  The data collection lasted approximately $1.5$~hours and produced $3{,}000{,}000$ synchronized pairs $(\mathbf{s}_k,\mathbf{x}_k,\mathbf{u}_k)$ for reconstruction training and evaluation.

\subsection{Reconstruction Model Training}
\label{subsec:model_training}

The synchronized dataset is processed according to the reconstruction formulation in Sec.~\ref{sec:framework}. The motion-capture poses are expressed relative to the robot base frame and converted to the candidate shape-state representations defined in Sec.~\ref{subsec:shape_state}. Similarly, each IMU orientation is expressed relative to the base IMU, and the resulting IMU and actuator measurements are assembled into the sensing configurations defined in Sec.~\ref{subsec:sensing_inputs}.

For each combination of shape-state representation, network architecture, training objective, and sensing inputs, multiple hyperparameter configurations (omitted for brevity) are trained and evaluated. In total, more than $5{,}000$ models are trained across the one- and three-segment systems. After training, the reconstructed states $\hat{\mathbf{x}}_k$ are converted to geometric body positions, allowing all candidate models to be compared using the common reconstruction metrics introduced in Sec.~\ref{subsec:shape_error}.

\subsection{Shape Reconstruction Metrics}
\label{subsec:shape_error}

Reconstruction performance is evaluated using three complementary geometric metrics: tip error, global shape error, and local shape error. Let $\mathbf{p}_{i}^{\{F\}}$ and $\hat{\mathbf{p}}_{i}^{\{F\}}$ denote the ground-truth and reconstructed positions of the $i$-th body frame expressed in the robot base frame $\{F\}$, respectively. The tip error measures the Euclidean reconstruction error of the distal body frame,
\begin{equation}
    e_{\mathrm{tip}}
    =
    \left\|
        \hat{\mathbf{p}}_{N_b}^{\{F\}}
        -
        \mathbf{p}_{N_b}^{\{F\}}
    \right\|_2 .
    \label{eq:tip_error}
\end{equation}
Although $e_{\mathrm{tip}}$ provides an intuitive measure of distal accuracy, it does not characterize reconstruction errors distributed along the robot body. We therefore additionally adopt the global and local shape errors introduced in \cite{wang_real-time_2026},
\begin{equation}
    e_{\mathrm{global}}
    =
    \frac{1}{N_b-1}
    \sum_{i=2}^{N_b}
    \left\|
        \hat{\mathbf{p}}_{i}^{\{F\}}
        -
        \mathbf{p}_{i}^{\{F\}}
    \right\|_2 ,
    \label{eq:global_shape_error}
\end{equation}
and
\begin{equation}
    e_{\mathrm{local}}
    =
    \frac{1}{N_b-1}
    \sum_{j=1}^{N_s}
    \sum_{i\in\mathcal{I}_j}
    \left\|
        \hat{\mathbf{p}}_{i}^{\{B_j\}}
        -
        \mathbf{p}_{i}^{\{B_j\}}
    \right\|_2 ,
    \label{eq:local_shape_error}
\end{equation}
where $N_s$ denotes the number of robot segments, $\mathcal{I}_j$ denotes the set of body frames belonging to segment $j$, and $\{B_j\}$ denotes the corresponding segment-base frame. The global error $e_{\mathrm{global}}$ characterizes the overall robot configuration, whereas $e_{\mathrm{local}}$ characterizes deformation within individual segments. Since the three metrics exhibit consistent trends in the subsequent ablation study, $e_{\mathrm{global}}$ is used as the representative full-body reconstruction metric in Fig.~\ref{fig:reconstruction_ablation}.

\subsection{Reconstruction Design and Ablation Study}
\label{subsec:model_selection}

\begin{figure}[t]
    \centering
    \includegraphics[width=0.95\linewidth]{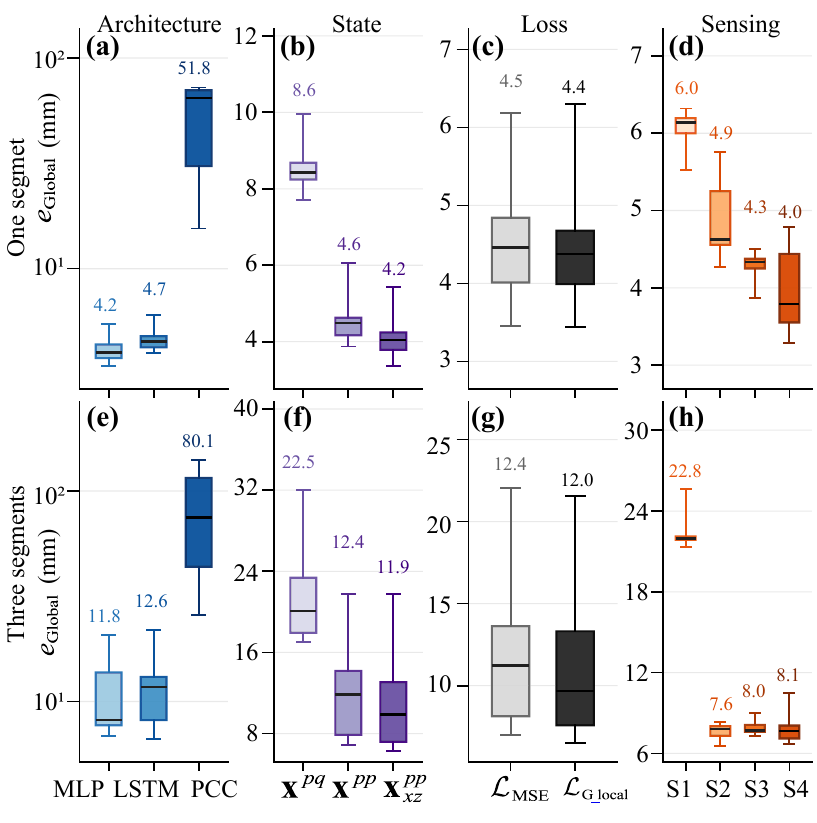}
    \caption{Ablation study of embedded full-body shape reconstruction using $e_{\mathrm{global}}$ for the one-segment (a)--(d) and three-segment (e)--(h) robots. The four columns evaluate (a),(e) network architecture, (b),(f) shape-state representation, (c),(g) training objective, and (d),(h) sensing inputs. Values above the box plots denote the mean global shape error in millimeters. S1--S4 denote the increasing sensing configurations: S1: encoder; S2: encoder + IMU; S3: encoder + IMU + torque; and S4: encoder + IMU + torque + input ($\mathbf{u}_{k-1}$).
    }
    \label{fig:reconstruction_ablation}
\end{figure}

Using the training pipeline above, we systematically evaluate the four reconstruction design choices introduced in Secs.~\ref{subsec:shape_state}--\ref{subsec:sensing_inputs}: shape-state representation, network architecture, training objective, and sensing inputs. Fig.~\ref{fig:reconstruction_ablation} reports the  $e_{\mathrm{global}}$ as the representative metric, while $e_{\mathrm{tip}}$ and $e_{\mathrm{local}}$ exhibit similar trends and are omitted for brevity.

Several consistent design trends are observed across both robot configurations. First, the learned MLP and LSTM estimators substantially outperform the PCC reconstruction, achieving an approximately $89\%$ reduction in mean reconstruction error. Among the learned architectures, the MLP achieves slightly lower reconstruction error than the LSTM for the operating regime considered here.

Second, the projected global--local representation $\mathbf{x}^{pp}_{xz}$ achieves the lowest reconstruction error among the candidate shape states, providing an average improvement of approximately 51.5$\%$. This result is consistent with \cite{ristich_shape_2025}, where removing explicit orientation states and the weakly modeled extension direction provides a more feasible representation for data-driven modeling. Denser shape states provide lower reconstruction errors; these results are omitted for brevity.

Third, the global--local training objective $\mathcal{L}_{\mathrm{glocal}}$ provides a modest improvement of approximately $3.6\%$ over conventional MSE. Although smaller than the effects of architecture and sensing inputs, this result indicates that explicitly considering both global configuration and segment-local deformation can improve full-body reconstruction.

Finally, the sensing ablation shows that adding distributed IMU measurements to encoder feedback provides substantial improvement for both robots. For the three-segment robot, the $e_{\mathrm{global}}$ decreases from $22.8~\mathrm{mm}$ with encoder feedback alone (S1) to $7.6~\mathrm{mm}$ with encoder and IMU measurements (S2), corresponding to a $67\%$ reduction. A similar improvement is observed for the one-segment robot. In contrast, torque and input information (S3--S4) further improve the one-segment reconstruction but provide no additional benefit for the three-segment robot. The contribution of these sensing measurements requires further investigation.

\subsection{Selected Online Reconstruction Pipeline}
\label{subsec:online_pipeline}

Based on the ablation study, the projected global--local state $\mathbf{x}^{pp}_{xz}$ and the global--local training objective $\mathcal{L}_{\mathrm{glocal}}$ are selected for the subsequent control experiments. The selected MLP contains two hidden layers with $256$ ReLU units per layer and is trained with a learning rate of $10^{-3}$ without dropout. The selected LSTM contains two recurrent layers with a hidden dimension of $256$, a sequence length of $H=100$, a stride of $50$, and a dropout rate of $0.1$, and is trained with the same learning rate. Both reconstruction models operate on embedded measurements sampled at $400~\mathrm{Hz}$. Longer temporal histories are also evaluated for the LSTM; however, increasing the sequence length to $H=200$ prevents the complete online control loop from maintaining the target $400~\mathrm{Hz}$ update rate. We therefore select $H=100$ as a trade-off between temporal information and real-time computation.

After training and model selection, the reconstruction networks are frozen and external motion capture is removed from the online pipeline. During operation, IMU and actuator feedback are processed by the frozen network to generate the estimated full-body shape state $\hat{\mathbf{x}}_k$ at $400~\mathrm{Hz}$. The reconstructed state is then used for Koopman operator identification and closed-loop control in Sec.~\ref{subsec:koopman_control}.

\subsection{Shape Reconstruction Performance}
\label{subsec:reconstruction_results}

\begin{figure}[t]
    \centering
    \includegraphics[width=\linewidth]{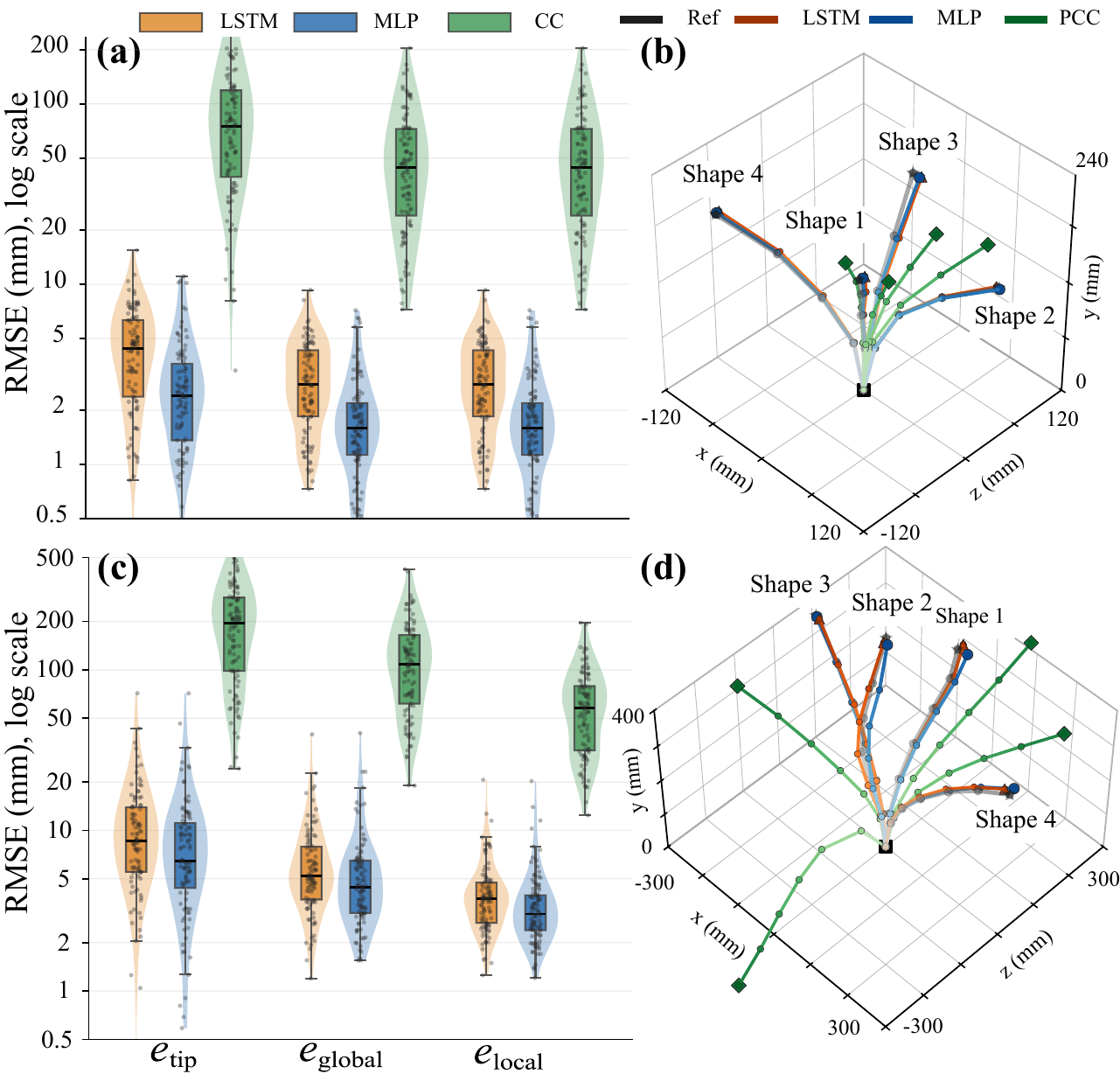}
    \caption{
    Embedded full-body shape reconstruction on the one- and three-segment continuum robots. (a), (c) Reconstruction error distributions for the MLP, LSTM, and CC estimators on the one- and three-segment systems, respectively. (b), (d) Representative reconstructed configurations compared with the motion-capture reference. 
    }
    \label{fig:shape_reconstruction_results}
\end{figure}

Fig.~\ref{fig:shape_reconstruction_results} compares the selected NN and PCC with ground truth for both the one- and three-segment robots. For the one-segment robot, both learned NN reconstruct the full-body shape with global shape errors below approximately 0.9$\%$ of the robot length, and substantially outperform the PCC baseline (Fig.~\ref{fig:shape_reconstruction_results}(a)). Four randomly selected configurations are shown in Fig.~\ref{fig:shape_reconstruction_results}(b), where the MLP and LSTM reconstructions closely overlap the reference shapes, whereas the PCC estimates exhibit substantial geometric deviations. The same trend becomes more pronounced for the three-segment robot (Fig.~\ref{fig:shape_reconstruction_results}(c, d)). These results indicate that the physical robot does not strictly follow the assumed PCC geometry, which would require additional distributed measurements.

Among the learned reconstruction models, the MLP achieves slightly lower reconstruction errors than the LSTM for both robot configurations. The experiments considered here primarily cover relatively low-speed motions with repeatable robot behavior, for which the instantaneous embedded measurements already contain sufficient information to reconstruct the shape accurately. 

The following experiments therefore evaluate whether this reconstruction accuracy is sufficient to replace external motion capture during real-time tip and full-body shape control.
\section{Control Experimental Results}
\label{sec:experiment}
This section extends the evaluation beyond shape estimation to physical closed-loop control. We evaluate reconstruction on one-segment (178~mm) and three-segment (516~mm) continuum robots, followed by one-segment tip tracking, three-segment full-body shape control, and enclosed-space operation without external observation.

\subsection{One-Segment Tip Trajectory Control}
\label{subsec:tip_control}

\begin{figure}[t]
    \centering
    \includegraphics[width=\linewidth]{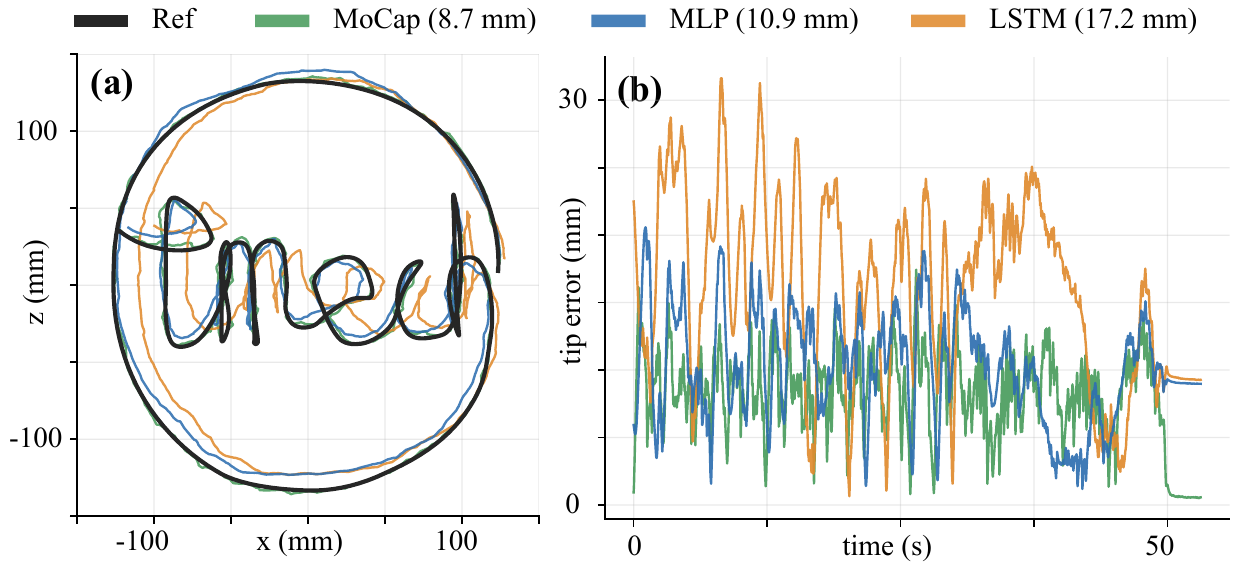}
    \caption{
     One-segment tip tracking of a ``tmech'' reference trajectory using external and embedded shape feedback.
(a) Tip trajectories using MoCap, MLP, and LSTM feedback.
(b) Corresponding tip tracking errors over time.
    }
    \label{fig:tip_control_results}
\end{figure}

We first evaluate whether the reconstructed shape state can replace external motion capture for one-segment tip trajectory tracking. Fig.~\ref{fig:tip_control_results} compares closed-loop tracking using direct MoCap, MLP, and LSTM feedback, achieving mean tip errors of $8.7~\mathrm{mm}$ ($4.9\%L$), $10.9~\mathrm{mm}$ ($6.1\%L$), and $17.2~\mathrm{mm}$ ($9.7\%L$), respectively. Despite the additional reconstruction error, both embedded estimators successfully track the reference trajectory without external observation, with the MLP achieving performance close to direct MoCap feedback. The lower MLP tracking error is also consistent with its higher reconstruction accuracy reported in Sec.~\ref{subsec:reconstruction_results}. As shown in Fig.~\ref{fig:tip_control_results}(a), the offsets reflect residual reconstruction errors propagated into the feedback state. After the trajectory ends at $50~\mathrm{s}$, Fig.~\ref{fig:tip_control_results}(b) further shows that MoCap converges to the smallest steady-state error, while the MLP and LSTM retain residual offsets associated with reconstruction bias.

This result highlights the direct coupling between reconstruction accuracy and closed-loop control performance: reconstruction bias propagates into the feedback state and consequently appears as a tracking offset. Nevertheless, the embedded shape estimates remain sufficiently accurate to enable closed-loop tip trajectory tracking.

\begin{figure*}[t!]
    \centering
    \includegraphics[width=0.9\linewidth]{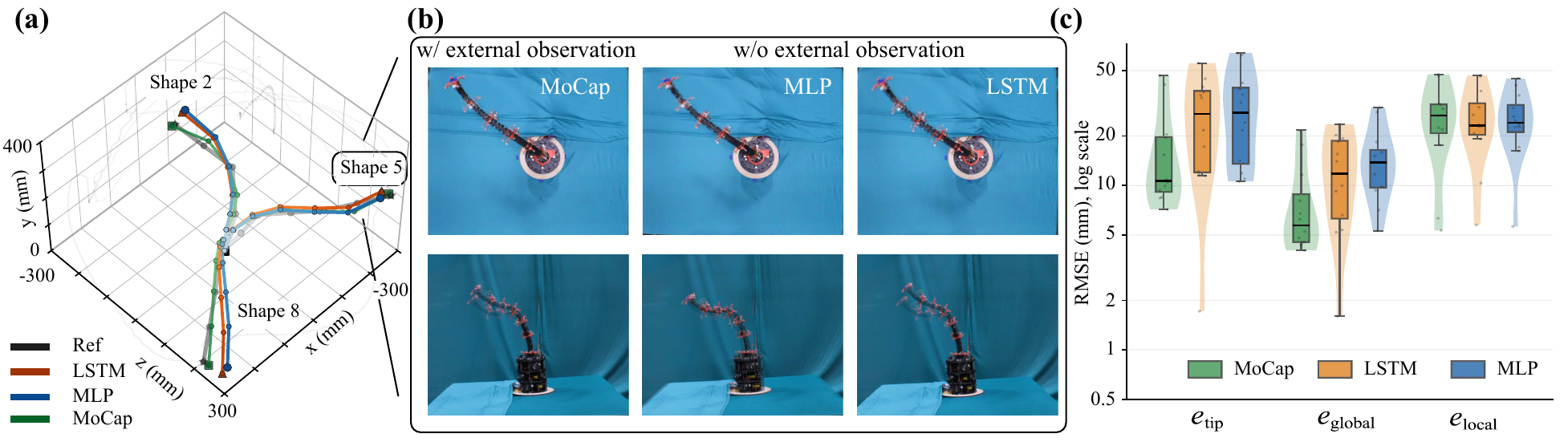}
    \caption{
    Three-segment full-body shape control using external and embedded shape feedback.
    (a) Representative target shapes and steady-state robot configurations obtained using MoCap, MLP, and LSTM feedback.
    (b) Front and top views of the controlled robot for Shape~5. The target shape is shown as a semi-transparent overlay on the measured robot configuration.
    (c) Tip, global, and local shape-control RMSE over 10 randomly selected target shapes. 
    }
    \label{fig:shape_control_results}
\end{figure*}

\subsection{Three-Segment Shape Control}
\label{subsec:shape_control}

The estimated shape state is further evaluated on the three-segment shape control, rather than only the tip, as shown in Fig.~\ref{fig:shape_control_results}. The corresponding front and top views for Shape~5 in Fig.~\ref{fig:shape_control_results}(b) further illustrate that the proposed framework reproduces a full-body configuration successfully. The complete shape tracking is provided in the accompanying video.

To evaluate this performance across different configurations, we randomly select $10$ target shapes and report the steady-state tip, global, and local shape errors in Fig.~\ref{fig:shape_control_results}(c). As expected, direct MoCap feedback achieves the lowest tip and global errors, while MLP and LSTM feedback introduce additional errors due to uncertainty in the reconstructed shape state. These results demonstrate that the embedded shape sensing preserves sufficient global and local shape information to serve directly as feedback for multi-segment closed-loop control.

Interestingly, although the MLP achieves lower errors than the LSTM in Sec.~\ref{subsec:reconstruction_results}, the LSTM achieves lower control error for the shape-control metrics in Fig.~\ref{fig:shape_control_results}(c). This suggests that minimizing reconstruction RMSE alone may not necessarily produce the best feedback state for control, since temporal consistency and the characteristics of the reconstruction error can also affect system identification and closed-loop performance. The relationship between reconstruction accuracy and control performance requires further investigation.

\subsection{Enclosed-Space Demonstration}
\label{subsec:demonstration}

\begin{figure}[t]
    \centering
    \includegraphics[width=\linewidth]{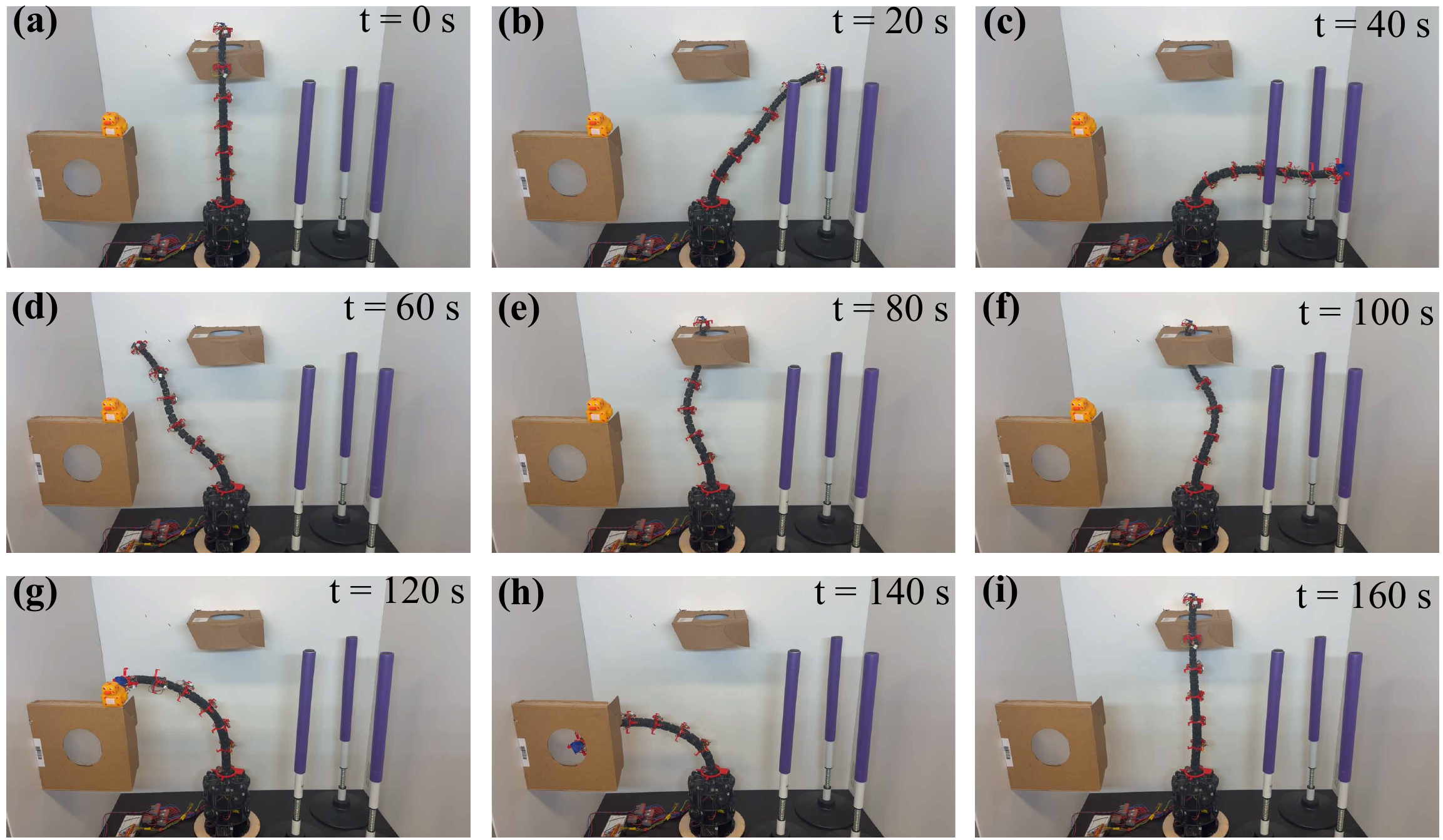}
    \caption{Enclosed-space demonstrations using the proposed embedded shape sensing and control framework without external observation.
    (a)--(c) Obstacle slalom, 0--60~s, (d)--(f) confined-aperture traversal, 60--120~s, and
    (g)--(i) tip targeting, 120--180~s.
     }
    \label{fig:confined_demo}
\end{figure}

To further demonstrate the practical applicability of embedded shape sensing, we perform three enclosed-space tasks without external observation (180~s in total), as shown in Fig.~\ref{fig:confined_demo}. The robot tracks a pre-programmed shape trajectory to navigate through a sequence of openings in an enclosed structure, where continuous line-of-sight for external motion capture is unavailable. The tasks include an obstacle slalom, ~\ref{fig:confined_demo}(a)--(c), confined aperture traversal ~\ref{fig:confined_demo}(d)--(e), and target interaction (with very limited contact) ~\ref{fig:confined_demo}(f)--(g). Due to imperfect tracking, unintended collisions with the surrounding structure occur. However, these disturbances do not destabilize the controller, and the robot continues to track the reference shapes and successfully passes through all three tasks. This demonstration highlights the potential of the embedded shape sensing and control framework for continuum robots in enclosed and visually occluded environments.

\subsection{Discussion and Limitations}
\label{subsec:discussion}

The present experiments demonstrate embedded shape feedback primarily under slow motions. The reconstructed shape state is validated only within the Koopman-based modeling and control framework considered in this work. The current models are trained and evaluated without external loading or physical interaction, which may alter the relationship between embedded measurements and robot shape. Finally, the reference shapes are sampled from physically measured robot configurations rather than generated online. Extending the framework to faster dynamics, other control architectures, external loading, feasible reference generation, and ultimately fully autonomous operation remains important future work.


\section{Conclusion} 
\label{sec:conclusion}

In this work, we extended embedded shape sensing to embedded shape control for continuum robots. By combining readily available IMU with actuator feedback, the proposed shape reconstruction achieved global shape errors of $0.9\% L$ and $1.7\%L$ for the one- and three-segment systems, respectively. Closed-loop experiments achieved a $6.1\%L$ tip-tracking error for the one-segment robot and a $2.3\%L$ global shape error for the three-segment robot, compared with $4.9\%L$ and $1.2\%L$ using MoCap feedback, respectively. These results demonstrated a practical embedded shape-sensing and control framework for continuum-robot operation in enclosed spaces.



\bibliographystyle{IEEEtran}
\bibliography{refs}
\vspace{-6.0em}

\begingroup
\raggedbottom

\begin{IEEEbiography}[{\includegraphics[width=1in,height=1.25in,clip,keepaspectratio]{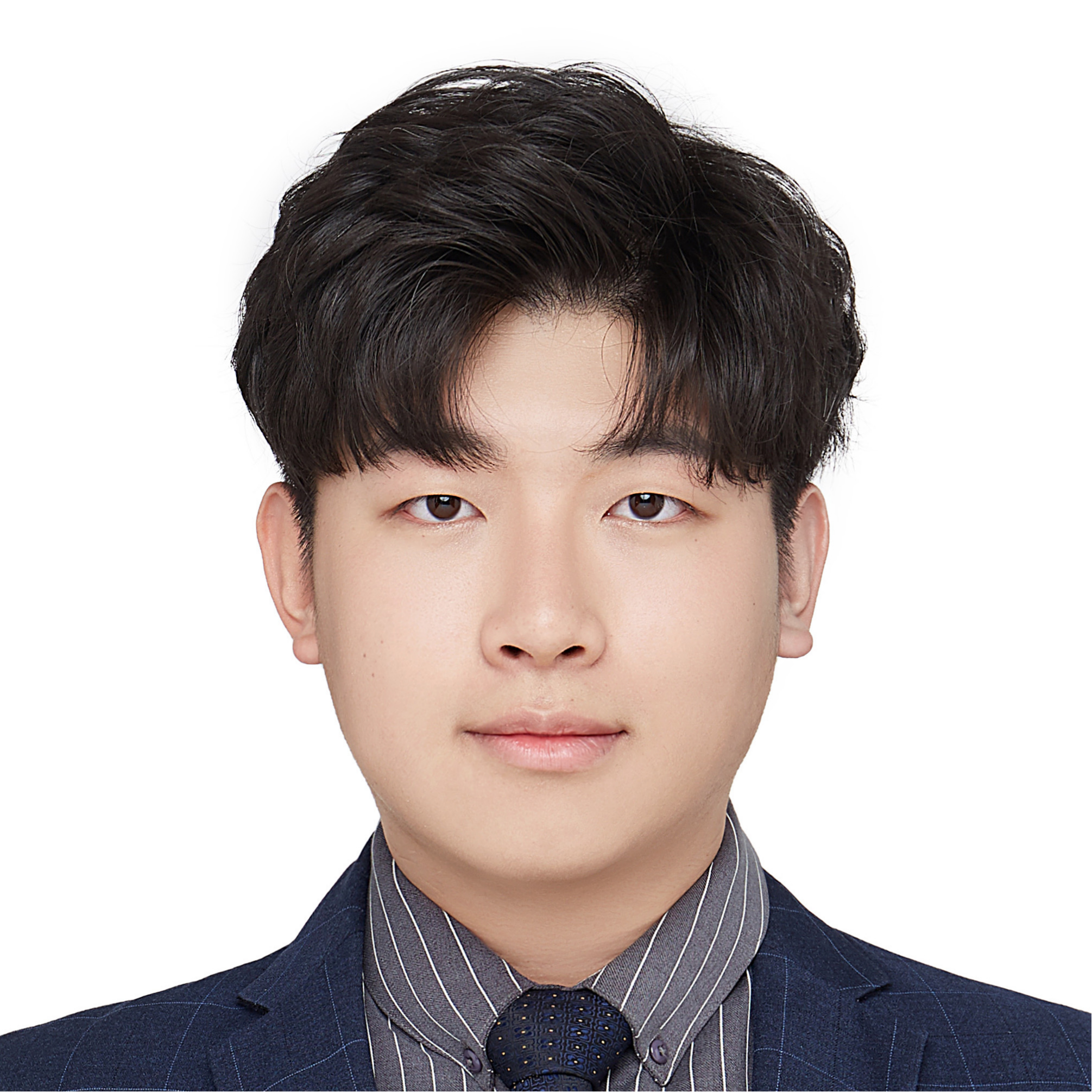}}]{Jiahe Wang} received the B.S. and M.S. degrees in mechanical engineering from Dalian University of Technology, Dalian, China, in 2021 and 2024, respectively. He is currently pursuing his Ph.D. degree in mechanical engineering at Arizona State University, Tempe, AZ, USA.

His research interests focus on data-driven modeling and control on soft robotics.
\end{IEEEbiography}
\vspace{-6.0em}

\begin{IEEEbiography}[{\includegraphics[width=1in,height=1.25in,clip,keepaspectratio]{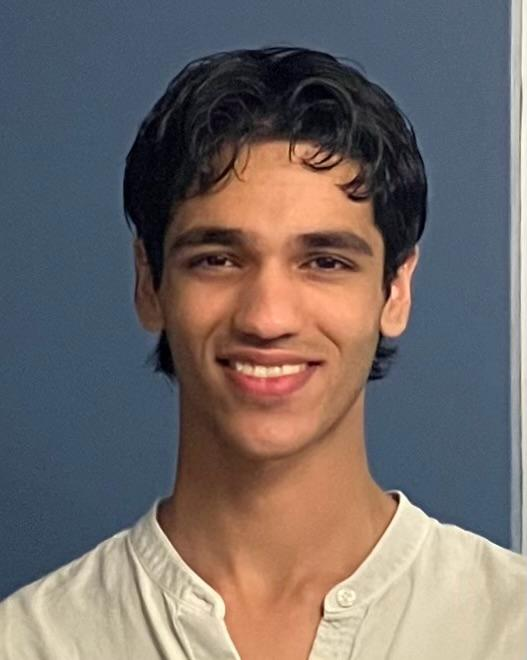}}]{Sultan Haidar Ali} is currently a junior undergraduate in Engineering (Robotics) at Arizona State University, Tempe, AZ, USA. He is a research assistant in the Robotics Actuators and Dynamics Laboratory at Arizona State University. 

His research interests include robot perception and state estimation.

\end{IEEEbiography}
\vspace{-6.0em}

\begin{IEEEbiography}[{\includegraphics[width=1in,height=1.25in,clip,keepaspectratio]{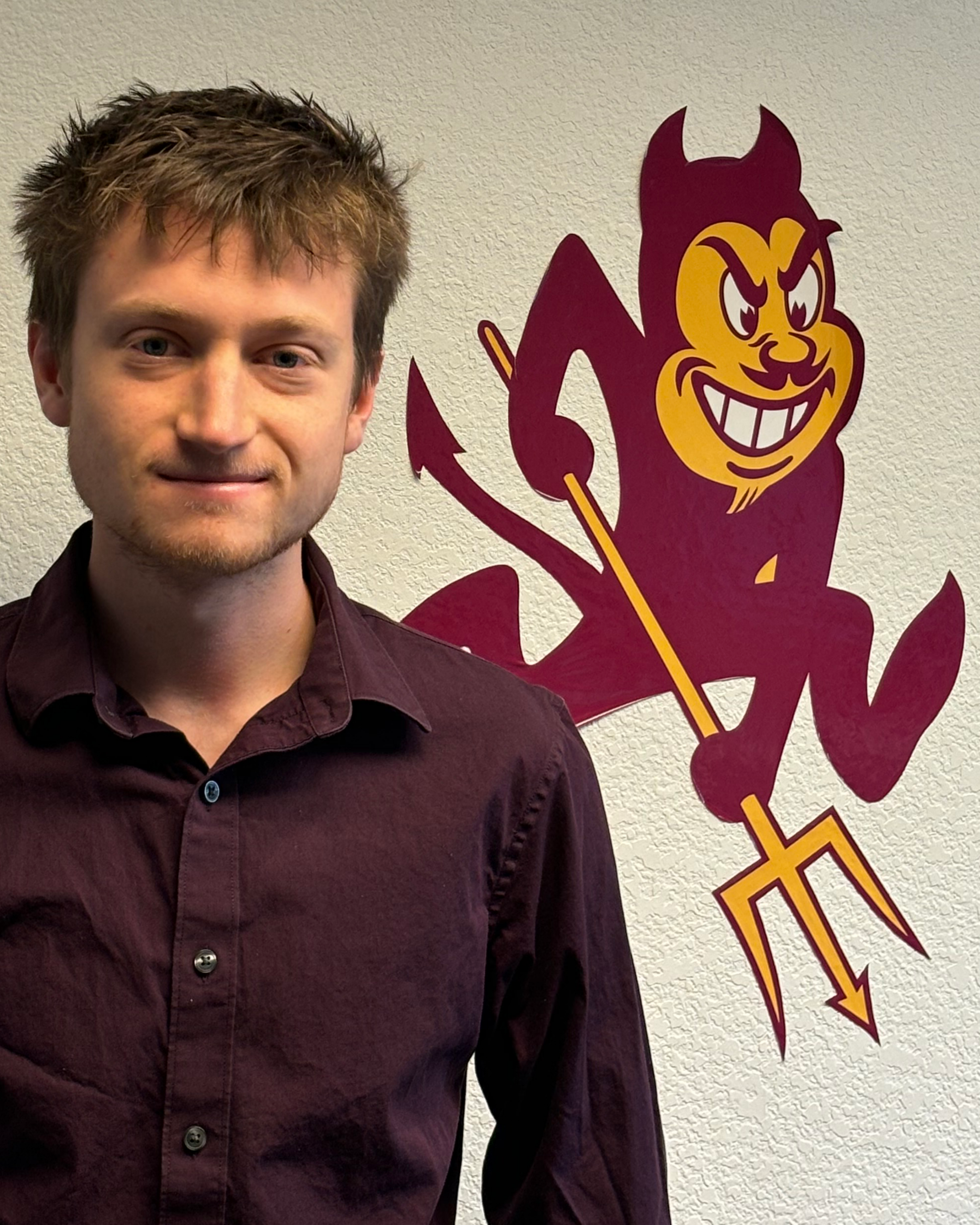}}]{Eric Weissman}
earned his B.S. in Aerospace Engineering from Arizona State University, Tempe, AZ, USA, in 2024 and is currently pursuing his Ph.D. in Mechanical Engineering at the same institution.

His research interests focus on soft robotic actuators and compliant robotics.
\end{IEEEbiography}

\vspace{-6.0em}

\begin{IEEEbiography}[{\includegraphics[width=1in,height=1.25in,clip,keepaspectratio]{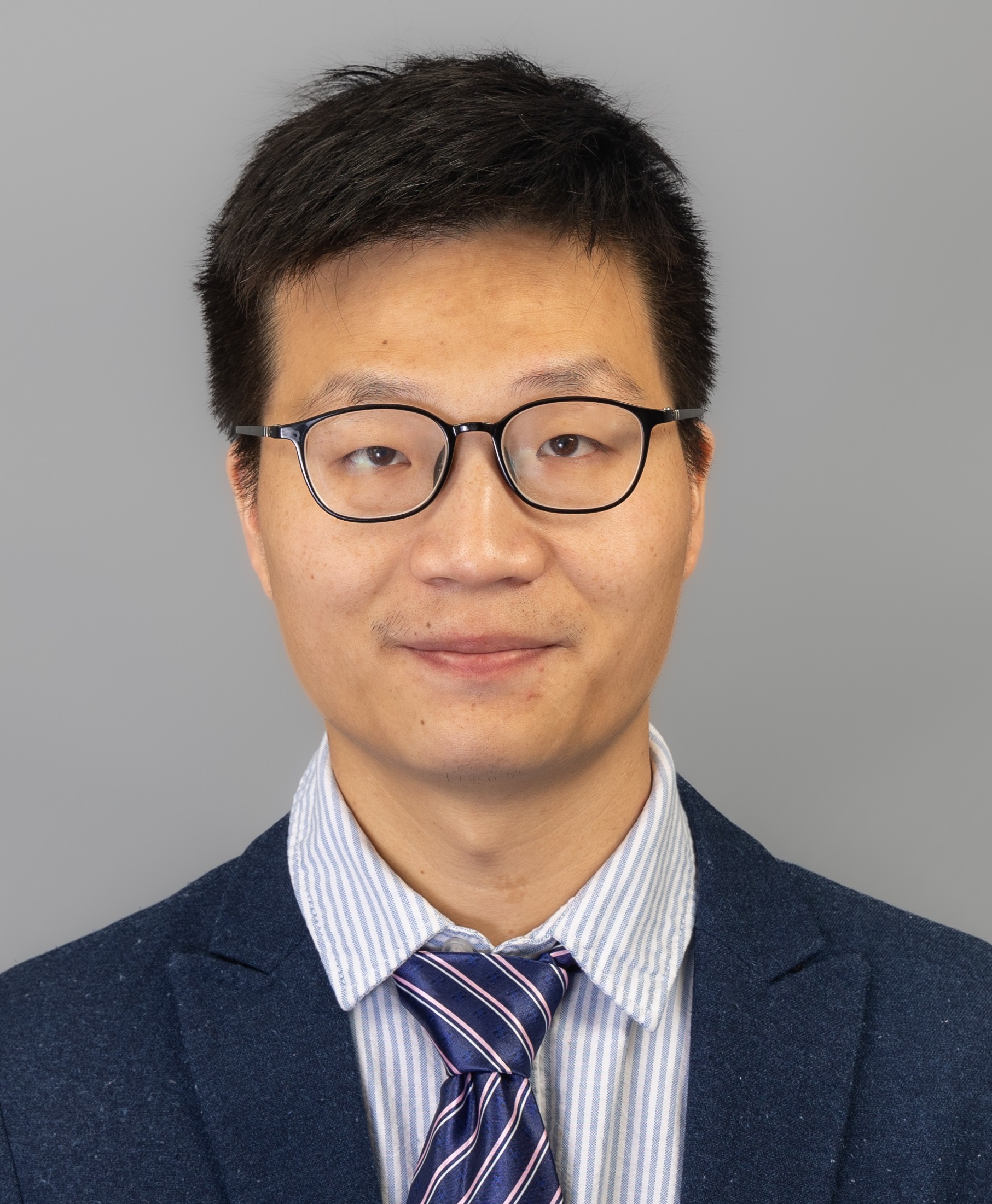}}]{Jiefeng Sun}
(Member, IEEE) received the B.S. degree in mechanical engineering from Lanzhou University of Technology, Lanzhou, China, in 2014, the M.S. degree in mechanical engineering from Dalian University of Technology, Dalian, China, in 2017, and the Ph.D. degree in mechanical engineering from Colorado State University, Fort Collins, CO, USA, in 2022.

He is currently an Assistant Professor at Arizona State University (ASU), Tempe, AZ, USA, and the Director of Robotic Actuators and Dynamics Lab at ASU. His current research interests include artificial muscle, soft robots, wearable robots, and reconfigurable robots.
\end{IEEEbiography}
\vspace{-6.0em}

\endgroup

\end{document}